\documentclass[runningheads]{llncs}
\usepackage{graphicx}
\usepackage{amsmath,amssymb} 
\usepackage{color}

\usepackage{epsfig}
\usepackage{latexsym, bm}
\usepackage{algorithm}
\usepackage{xspace}

\newcommand{\etal}{\textit{et~al.}}

\newcommand{\ie}{\textit{i}.\textit{e}.}
\newcommand{\eg}{\textit{e}.\textit{g}.}
\newcommand{\cadik}{{\^C}ad{\'\i}k~}

\newcommand{\Fref}[1]{Figure~\ref{#1}}

\newcommand{\eref}[1]{Eq.~(\ref{#1})}
\newcommand{\fref}[1]{Fig.~\ref{#1}}

\begin{document}

\newcommand{\point}{
    \raise0.7ex\hbox{.}
    }


\pagestyle{headings}

\mainmatter

\title{Perceptually Consistent\\ Color-to-Gray Image Conversion} 

\titlerunning{Perceptually Consistent\\ Color-to-Gray Image Conversion} 

\authorrunning{Shaodi You~~~~Nick Barnes~~~~Janine Walker} 

\author{Shaodi You, Nick Barnes, Janine Walker} 
\institute{National ICT Australia (NICTA), Australian National University} 

\maketitle

\begin{abstract}
	In this paper, we propose a color to grayscale image conversion algorithm (C2G) that aims to preserve the perceptual properties of the color image as much as possible.
	To this end, we propose measures for two perceptual properties based on contemporary research in vision science: brightness and multi-scale contrast. 
	The brightness measurement is based on the idea that the brightness of a grayscale image will affect the perception of the probability of color information.
	The color contrast measurement is based on the idea that the contrast of a given pixel to its surroundings can be measured as a linear combination of color contrast at different scales.
	Based on these measures we propose a graph based optimization framework to balance the brightness and contrast measurements.
	To solve the optimization, an $\ell_1$-norm based method is provided which converts color discontinuities to brightness discontinuities.
	To validate our methods, we evaluate against the existing \cadik and Color250 datasets, and against NeoColor, a new dataset that improves over existing C2G datasets.
	NeoColor contains around 300 images from typical C2G scenarios, including: commercial photograph, printing, books, magazines, masterpiece artworks and computer designed graphics.
	We show improvements in metrics of performance, and further through a user study, we validate the performance of both the algorithm and the metric.
\end{abstract}

\section{Introduction}

Color images are frequently presented in grayscale.
This is common with digital ink displays and grayscale-only printing,
but synthesizing a consistent grayscale image is also important for color blindness, and for prosthetic vision devices such as retinal implants\cite{AytonPLOSONE14} and sensory substitution devices \cite{Stronks2015} which mostly convey brightness only.
One may also wish to present color information in grayscale as part of a virtual or augmented reality application.
Such conversion is sometimes considered in image processing tasks such as tone-mapping and compression.
For these reasons converting color images to grayscale images (C2G) has attracted a stream of research attention (\eg, \cite{Bala03,Gooch05,Yoo15}).


Despite its common usage, C2G is non-trivial. 
Normal human vision perceives color with three peak spectral sensitivities \cite{Pokorny2004}.
Hence, perceivable color-space is homogeneous to $\mathbb{R}^3$, whereas gray-scale space is homogeneous to $\mathbb{R}$. 
Standard displays present an image in a smaller discrete range though dynamic range and quantization restrictions.
Perception is restricted further by human observers' ability to perceive brightness and color difference \cite{Stevens57}.
However, as perception of the presented grayscale image is also restricted, it is inevitable that information loss will occur in C2G.

\begin{figure}[tb]
	\includegraphics[width=\linewidth]{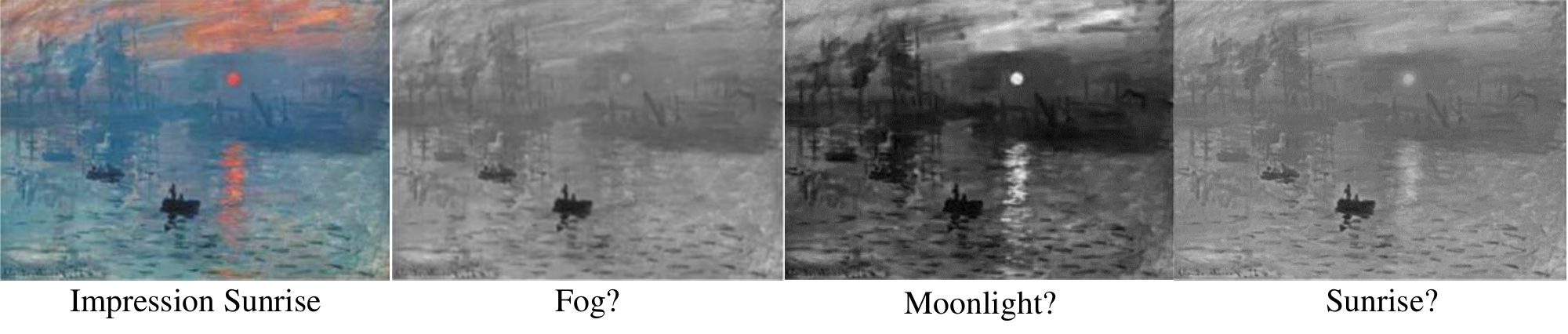}
	\vspace{-20pt}
	\caption{Converting Claude Monet's masterpiece "Impression Sunrise" to grayscale.} A conversion without considering the color contrast will change the feeling to haze or fog while a conversion without considering the brightness will change the feeling to moonlight. 
	\label{Fig:Teaser}
	\vspace{-12pt}
\end{figure}

More importantly, grayscale space has strong perceptual meaning. 
Simply quantizing the 3D signal and presenting it as a 1D signal often will not lead to a grayscale image that is perceptually consistent with the original color image .
An example is shown in \fref{Fig:Teaser}, Claude Monet's masterpiece "Impression Sunrise", which is included in \cadik dataset \cite{Cadik08} and has been actively used for validating C2G algorithms. 
As can be seen, while the original color image gives the feeling of sunrise, that feeling can easily be lost in conversion to grayscale.
A direct grayscale conversion without considering color contrast can change the feeling of the image to fog or haze.
On the other hand, strongly enhancing contrast while allowing the brightness to change aggressively can change the feeling of the image to moonlight.
Further, some representations of color can lead to an impression of the scene where the appearance is not natural.
For example, consider the first row of \fref{Fig:VisualNeo}, where as a result of methods that emphasize contrast over perceptual consistency, orange juice may appear black.

Another important consideration is that "contrast" is a relative concept that is highly reliant on context. A famous example is the shadow illusion by Edward Adelson \etal\cite{Adelson00}.
While earlier research in C2G focused more on contrast in color space, current research considers spatial context. However, perceiving spatial context is rather complex. Thus, a perceptional consistent contrast model is desired.

In this paper, we propose a C2G algorithm which aims to preserve perceptual consistency as much as possible.
To this end, we propose two perceptual consistency measurements based on contemporary research in vision science: brightness and multi-scale contrast measurement.
The brightness measurement is based on the idea that brightness in a grayscale image does affect the perception of the probability of color information.
The color contrast measurement is based on the finding that the contrast of a given pixel to its surroundings can be measured as a linear combination of color contrast in different scales.
Based on these measures, we propose graphical model based optimization framework to balance the brightness and contrast measurement. To solve the optimization, an $\ell_1$-norm based method is provided.

To validate our methods, we also propose a new dataset, called NeoColor, that improves on existing datasets.
The \cadik dataset \cite{Cadik08} has only 24 images. Further, the Color 250 dataset \cite{Lu14} is a subset of Berkeley segmentation dataset which was not designed to evaluate C2G algorithms. For NeoColor, we collect about 300 images from typical C2G scenarios including: natural scenes from commercial printing, books, magazines, masterpiece artworks, and computer designed graphics.

The contribution of our work is three-fold.
Firstly, we propose a rationale for perceptual consistency of C2G algorithms. 
Based on perceptual consistency, we propose brightness and multi-scale color contrast measurements.
Secondly, we propose an efficient algorithm for C2G which best preserves the perceptual consistency measurement.
Thirdly, a new dataset which is specially designed for C2G evaluation is provided to address some shortcomings of the \cadik and Color250 datasets.

The rest of the paper is organized as follows: Section 2 summarizes related work on C2G. In Section 3, we discuss key aspects of human perception. 
Based on this, in Section 4, the perceptual consistency C2G algorithm is described. Section 5 demonstrates the effectiveness of the proposed method using both automatic evaluation and a user study. Section 6 concludes the paper.

\section{Related Work}

Existing C2G research can be categorized as global and local methods. Generally, global methods seek a global mapping for each color value to an intensity value, aiming to maximize the overall contrast, whereas local methods enhance spatially local contrast.

\vspace{-12pt}
\paragraph{Global methods:}
Bala \etal \cite{Bala03} start with an image with a limited number of colors, the colors are sorted and brightness values are assigned accordingly.
Gooch \etal \cite{Gooch05} considered brightness variation according to hue, the orientation of the hue-ring was determined by maximizing the global contrast.
Kim \etal \cite{Kim09} extended Gooch \etal's method: variation from the hue-ring is represented by a Fourier series instead of the original linear mapping.
Ancunti \etal \cite{Ancuti11} also extended the method of Gooch \etal, so that salient and non-salient areas have different hue-ring variation. A user study was also included to evaluate the effectiveness.

Rosche \etal \cite{Rasche05,Rasche05re} sought a continuous global mapping to assign similar colors to similar grayscale values, using pair-wise comparison between sampled colors.
Using the same framework, Grundland \etal \cite{Grundland07} improved the computational efficiency by using PCA to find the 1D component that maximizes contrast.

In more recent work, spatial context has been taken in consideration for global methods. Kuk \etal \cite{Kuk10} proposed method to find landmark colors though K-means. Adopting similar strategy for spatial context, Lu \etal \cite{Lu12,Lu14} proposed a bimodal contrast preserving energy function, along with local and non-local constraints. An evaluation metric and dataset were also introduced to automatically evaluate C2G algorithms. 
Ji \etal \cite{Ji15} proposed a hybrid method where color mapping is performance by modulating the hue ring and edges are sharpened by bandpass filtering.
Yoo \etal \cite{Yoo15} considered video C2G instead of single image C2G. They clustered the image sequence in spatio-temporal space to a 1D manifold, and assigned brightness accordingly.

\vspace{-12pt}
\paragraph{Local methods:}
Bala \etal \cite{Bala04} proposed high pass filtering the color image to strengthen local contrast when converting to grayscale.
Subsequently, Smith \etal \cite{Smith08} presented a method to pre-assign color variation based on vision science studies, and applied the variation to enhance local contrast.
In recent work, Liu \etal \cite{Liu15} proposed an enhancement which preserves gradient, while attenuating the halo between inconsistent boundaries and
local method of Du \etal\cite{Du15} emphasised color contrast in salient areas.

Existing methods focus on best preserving contrast. However, inadequate focus has been given to ensuring that the image is consistent with human perception.

\section{Human Perception of Color and Measurement}

Modeling human perception of color in an algorithm is not trivial. It relies not only on physical modeling of object luminance and spectrum, but also on studies of human visual perception.
Before we introduce our algorithm, which aims to produce a grayscale image that is consistent with human perception of the original color image, in this section, we discuss the factors that affect human color perception. 
We first introduce the basic concepts in color space and our notation. 
In Sec. 3.1 we discuss the perception of color brightness with a probability model, and introduce an energy function to model brightness perception.
Then, in Sec. 3.2 we discuss perception of color contrast in complex spatial contexts and introduce an energy function to model complex color contrast perception.
\\
\\
\textbf{Background and notation}
\vspace{-12pt}
\paragraph{Color space:} Since Young-Helmholz theory in the mid-1800s, it has been generally agreed that human perception of color is within a three dimensional space \cite{Campbell68,Lotto02,Peli90,Haun13,Stevens57,Majumder07,Jaroensri15}.
In images, this can be represented in several common color spaces, \eg, RGB, HSL, YUV, CIEL*a*b*.
Because CIE-LAB preserves the perceived difference between colors, in this paper, we will use Euclidean distance in L*a*b* space as the contrast metric, which is common practice for recent C2G research \cite{Lu14,Du15,Ma15,Liu15}.
\Fref{Fig:ColorRing}.a is an illustration of the CIEL*a*b* color space: a spherical space where lightness $l$ varies from 0 for black to 100 for white. $a$ varies from 100 for red to -100 for green; and $b$ varies from 100 for yellow to -100 for blue.

\vspace{-6pt}
\paragraph{Notation:} Notation is listed in Table \ref{T:Denotation}, for the convenience of the reader. Specifically, we denote the color channels of a pixel as $l,a,b$ respectively and the targeted grayscale value as $g$. We use lower case for a single pixel, capitalized letter for image matrix and bold lower case for vectors. 
\begin{table}[tb]
	\caption{Notation} 
	\vspace{-6pt}
	\includegraphics[width=\linewidth]{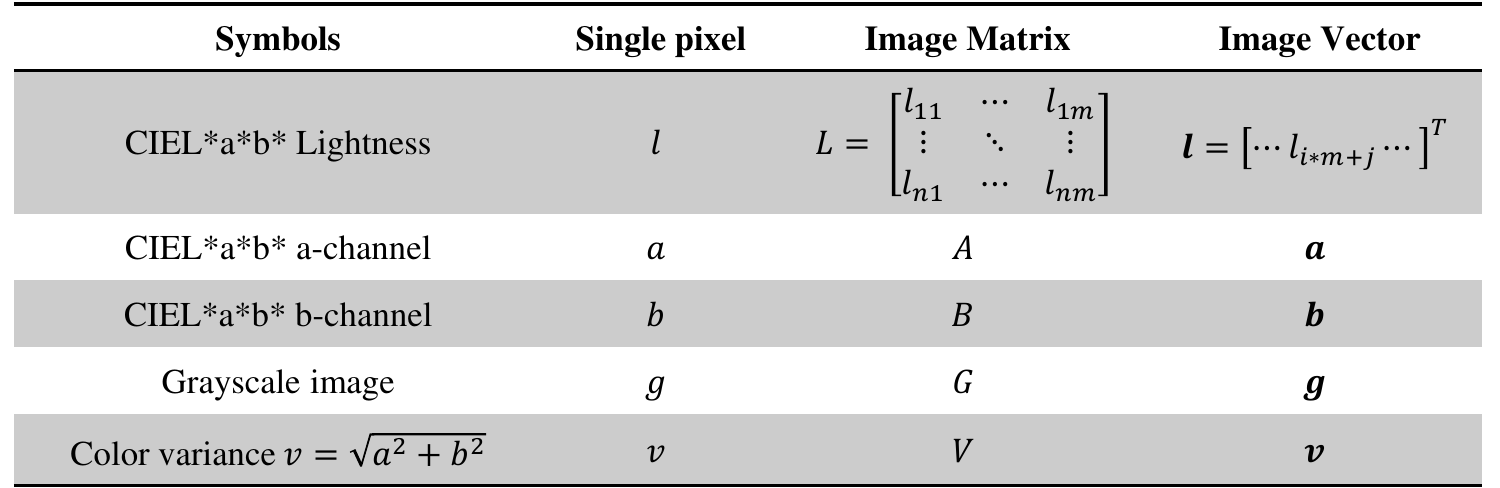}
	\label{T:Denotation}
	\vspace{-12pt}
\end{table}

\subsection{Perception of Color Brightness and Perception Measurement}

%

\paragraph{Color as function of brightness:}

\begin{figure}[tb]
	\includegraphics[width=\linewidth]{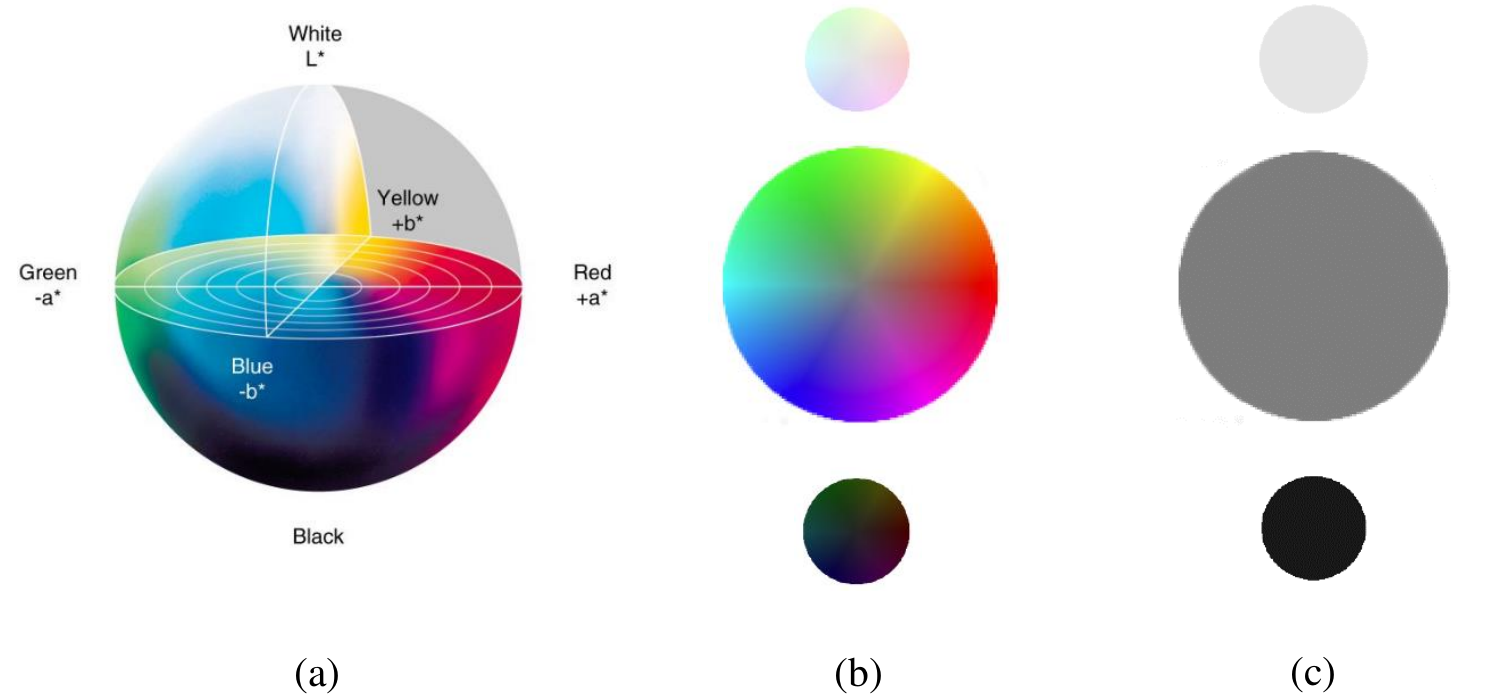}
	\vspace{-12pt}
	\caption{Color variance as function of brightness.} (a) The CIE L*a*b* color space. (b) A color slice in L*a*b* space with high, middle, low brightness. A high/low brightness color does not give same level of color contrast as mid-brightness color.
	(c) Desaturated slice of (b), a high/low brighness indicates a small color variation, while middle brightness indicates a large range for color variation.
	\label{Fig:ColorRing}
	\vspace{-12pt}
\end{figure}

As can be inferred from L*a*b* color space, color components with very low or very high brightness do not provide the same level of color variance as those with mid brightness levels. We show a simple example in \fref{Fig:ColorRing}.

Unlike previous C2G, especially global methods, \cite{Lu14,Grundland07}, where brightness is considered to be independent of color, we consider brightness as an importance cue for indicating color.
As we presented in \fref{Fig:ColorRing}, although brightness does not allow us to infer the $a,b$ value of a color, it does allow us to infer the possibility of color. This is important for generating a grayscale image that preserves the perceivable feeling of color.

Here we discuss the possibility of infereing color $a, b$ as function of brightness $l$. Using the metric in CIEL*a*b* space, a color $(l, a, b)$ with brightness $l \in [0, 100]$ has a color variance, denoted as $v$, limited in the spherical color space:
\vspace{-6pt}
\begin{equation}
	\vspace{-6pt}
	v = \parallel (a(l), b(l))\parallel_2 \leq \sqrt{100 ^2 - 4 * (l - 50) ^ 2},
	\label{Eq:BCM2}
\end{equation}
Assuming a uniform distribution of color in CIEL*a*b* space, the partial probability of $v$ by $l$ is:\footnote{We use proportion instead of equal mark and neglect a constant for normalization.}
\vspace{-6pt}
\begin{equation}
	\vspace{-6pt}
	\begin{split}
		p(v = \tilde{v} | l)  \propto \tilde{v}\sqrt{100 ^2 - 4 * (l - 50) ^ 2},
		\\
		0 \leq \tilde{v} \leq \sqrt{100 ^2 - 4 * (l - 50) ^ 2}.
	\end{split}
	\label{Eq:BCM2}
\end{equation}
and similarly:
\vspace{-12pt}
\begin{equation}
	\vspace{-0pt}
	\begin{split}
		p(l = \tilde{l} | v)  = \frac{2}{\sqrt{100^2 - v^2}},
		\\
		50 - \frac{\sqrt{100^2 - v^2}}{2} < \tilde{l} < + \frac{\sqrt{100^2 - v^2}}{2} .
	\end{split}
	\label{Eq:BCM3}
\end{equation}

As can be seen, a color with a large $v$ value is less likely to have a large variance in $l$, whereas a color with large $l$ is less likely to have a large variance in $v$. 

\vspace{-0pt}
\paragraph{Brightness Perceptual Energy}
Considering \eref{Eq:BCM2} and \eref{Eq:BCM3}, we define the brightness perceptual energy for inconsistent brightness between the color and grayscale image as:
\vspace{-6pt}
\begin{equation}
	\vspace{-6pt}
	\begin{split}
		E_B(g) &= w(l,a,b)\parallel l - g \parallel _ {\ell}
		\\
		& =  \frac{1}{\sqrt{100^2 - a^2 - b^2} + \epsilon}  \frac{1}{\sqrt{100^2 - (2l - 100)^2} + \epsilon}\parallel l - g \parallel _ {\ell},
	\end{split}
	\label{Eq:BCM1}
\end{equation}
where $\epsilon$ is a small constant regulator.
Generally, when the brightness in a color image $l$ is not mapped to same grayscale value $g$, error energy increases. The error energy is reweighted by $w$ which penalizes color with large variance $v$ (because high probablity color is not sensed); and color with high or low brightness $l$  (because low probablity of color is not sensed).

\vspace{-6pt}
\subsection{Perception of Multi-Scale Spatial Contrast and Measurement}
In the previous subsection, we assume that the two areas of color appear close together spatially, and color contrast is only evaluated between the two. 
Much of the existing C2G research is based on such an assumption. 
However, color contrast is not absolute, but depends on spatial context.
In this section, we start by considering pairwise color contrast and simple spatial context.
Specifically, we introduce the Campbell-Robson curve, \cite{Campbell68,Oliva06}.
Based on this, we introduce perception of complex spatial contrast, specifically considering recent research from vision science \cite{Peli90,Haun13}.
Finally, we study the perception complex spatial color contrast as it appears in many real images.

\begin{figure}[tb]
	\vspace{0pt}
	\includegraphics[width=\linewidth]{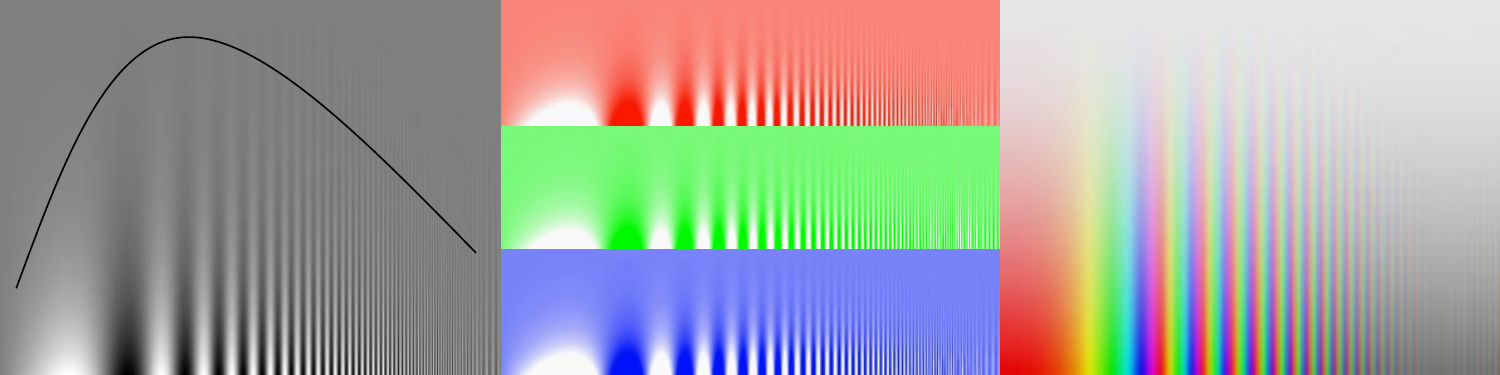}
	\caption{Campbell-Robson curve.} Left, the original Campbell-Robson curve, brightness contrast sensitivity is a function of spatial frequency. Middle, Campbell-Robson curve on single color. We assume color brightness contrast follows the same curve. Right, curve on hue ring, the color hue contrast follows the same curve.
	\label{Fig:Campbell}
	\vspace{-12pt}
\end{figure}

\vspace{-6pt}
\paragraph{Peception of simple spatial contrast:}

As illustrated in \fref{Fig:Campbell}.a, the Campbell-Robson curve demonstrates that the perception of contrast is a function of spatial frequency. Specifically, human perception of contrast is most sensitive at 1-2 cycles per degree (cpd). The sensitivity drops for spatial frequency above and below the peak.

The Campbell-Robson study is only of luminance contrast. No equivalent study is available for the perception of color contrast.
Although in practice, colour contrast is likely to exhibit spatial sensitivities that differ from luminance contrast, we take the simplest assumption that the perception of color contrast follows the same curve.
This is shown in \fref{Fig:Campbell}.b for the perceptional sensitivity of color-brightness contrast with respect to spatial frequency, and in \fref{Fig:Campbell}.c, for the perception of hue contrast.
Our subsequent algorithm could be adapted if curves like that Campbell-Robson became available for color contrasat.
Thus,

\noindent\textbf{Assumption 1} the perception of color contrast (Brightness, Hue, Saturation) with respect to spatial frequency follows the same Campbell-Robson curve.

\begin{figure}[tb]
	\vspace{-0pt}
	\includegraphics[width=\linewidth]{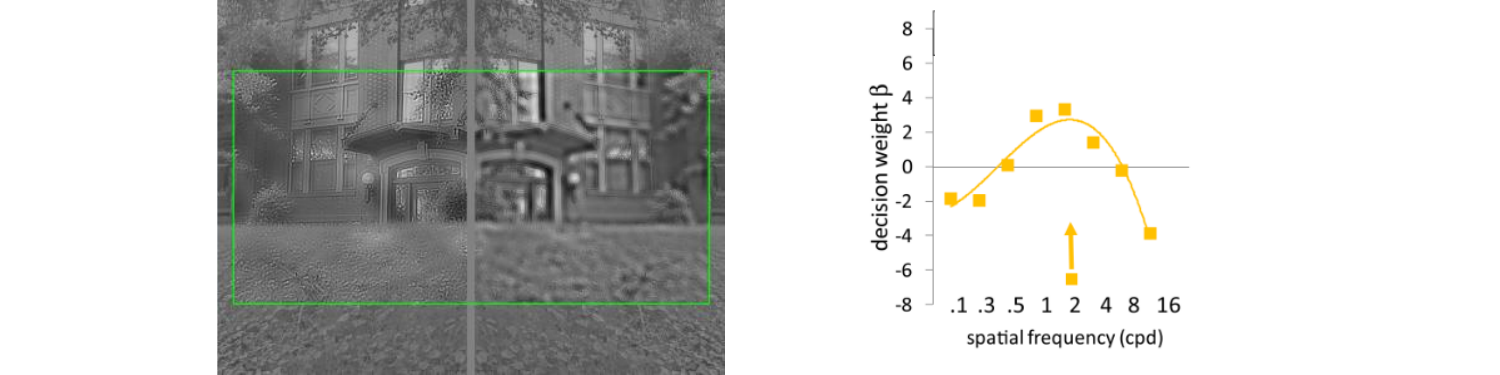}
	\caption{The Haun-Peli curve for perceiving complex contrast \cite{Haun13}. Left, two gray scale images with difference combinations of intensity of spatial frequency component. Users are asked to choose the one with higher contrast. Right, statistics of importance of different frequency components for the overall perceived contrast.
	} 
	\label{Fig:Haun}
	\vspace{-6pt}
\end{figure}

\begin{figure}[tb]
	\includegraphics[width=\linewidth]{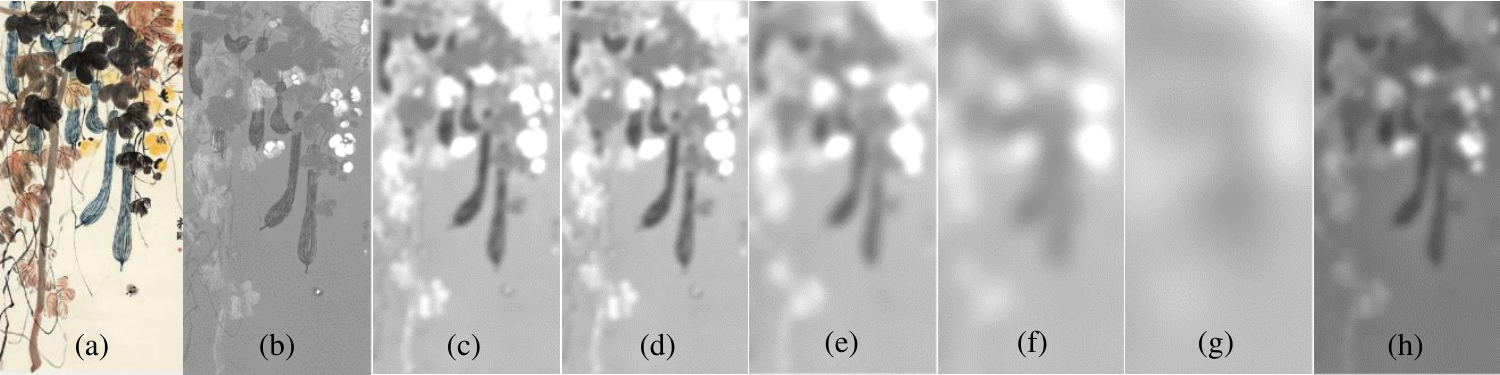}
	\vspace{-12pt}
	\caption{Complex color contrast. (a) Color image, (b) b channel of L*a*b* spece. (c - g) contrast in different scale, using DoG filter. (h) Combined complex contrast according to Haun-Peli curve.
	} 
	\label{Fig:Contrast}
	\vspace{-12pt}
\end{figure}

\vspace{-6pt}
\paragraph{Perception of complex spatial contrast:}
However, the spatial contrast of a natural image is often far more complex that the Campbell-Robson pattern. 
According to Haun \etal\cite{Haun13}, human perception of complex contrast follows a linear of combination of different spatial frequencies. 
In \fref{Fig:Haun}, we include a typical result of such a combination. Note that the coefficients for such a combination need not all be positive.

\vspace{-12pt}
\paragraph{Perception of complex spatial color contrast:}

Unfortunately, no systematic vision science model is available for complex color contrast. Therefore, here again, we make an assumption that the perception of complex brightness contrast can be extended to complex color contrast:

\noindent\textbf{Assumption 2} the perception of complex color contrast with respect to spatial frequency follows the same Haun-Peli curve.

\vspace{-12pt}
\paragraph{Complex Color Contrast Consistency Measurement:}

According to the above assumptions, we introduce the Brightness Contrast Measurement. For a single channel (grayscale) image the complex contrast is defined as:
\vspace{-6pt}
\begin{equation}
	\vspace{-6pt}
	c(G) = \sum\beta_i k_i * G,
	\label{Eq:CCM1}
\end{equation}
where $k_i$ is the Difference of Gaussian (DoG) Kernel at scale $2^i$:
\vspace{-6pt}
\begin{equation}
	\vspace{-6pt}
	k_i(j, k) = \frac{1}{2\pi 2^{2i}} e^{-\frac{j^2 + k^2}{2 ^{2i}}} - \frac{1}{2\pi 2^{2(i+1)}} e^{-\frac{1}{2}\frac{k^2 + k^2}{2^{2(i+1)}}}.
	\label{Eq:CCM2}
\end{equation}
$\beta_i$ are the coefficients derived from the Haun-Peli curve. For example, a screen with 72 dots per inch (dpi), and viewed from 60cm away.
\vspace{-6pt}
\begin{equation}
	\vspace{-6pt}
	\beta_1 = -4, \beta_2 = 1, \beta_3 = 4, \beta_4 = 4, \beta_5 = 1, \beta_6 = -2.
	\label{Eq:CCM3}
\end{equation}
When varying the dpi or the viewing distance, the scale by pixels shifts proportionally. \Fref{Fig:Contrast} shows an example of the contrast at different scales and the combined complex contrast.

We define the complex contrast for the color channels similarly and denote them as $c(L), c(A), c(B)$.  The contrast measure is defined as:
\vspace{-6pt}
\begin{equation}
	\vspace{-6pt}
	E_c(G) = \alpha_L\parallel c(L) - c(G) \parallel_{\ell} 
	+ \alpha_{AB}\parallel c(A) - c(G) \parallel_{\ell} 
	+ \alpha_{AB}\parallel c(B) - c(G) \parallel_{\ell},
	\label{Eq:CCM4}
\end{equation}
where $\alpha_L, \alpha_{AB}$ balances the importance between different channels.
As can be seen, the contrast energy functions has three terms that represent the contrast lost for the $l,a,b$ channels respectively. Note that the contrast loss for the $a,b$ channels are directional, this is because we incorporate the understanding of human vision that the brightness of same color is not always perceived in the same way \cite{Corney09}.

\section{Methodology}

In this section, we introduce a graphical model based method which aims to find a gray-scale image which best fits the brightness perceptual consistency as well as the contrast consistency. 

\vspace{-6pt}
\paragraph{Objective Function}
Combining \eref{Eq:BCM1} and \eref{Eq:CCM4}, the proposed objective function seeks the gray-scale image $G$ that optimizes the trade-off between brightness consistency and contrast:
\vspace{-6pt}
\begin{equation}
	\vspace{-6pt}
	E(G) = E_B(G) + E_C(G)
	\label{Eq:OBJ1}
\end{equation}

Similarly, \eref{Eq:OBJ1} can be represented as linear system:
\vspace{-6pt}
\begin{equation}
	\vspace{-6pt}
	\begin{split}
		E(\bm{g}) &= E_B(\bm{d}) + E_C(\bm{g})
		\\
		&= \parallel \Lambda(\bm{w})(\bm{l} - \bm{g})  \parallel_{\ell} 
		\\
		&~~~~+ \alpha_L\parallel C\bm{l} - C\bm{g} \parallel_{\ell} 
		+ \alpha_{AB}\parallel C\bm{a} - C\bm{g} \parallel_{\ell} 
		+ \alpha_{AB}\parallel C\bm{b} - C\bm{g} \parallel_{\ell},
	\end{split}
	\label{Eq:OBJ3}
\end{equation}
where the bold letters are the vectorized representation of the image matrix. $C$ is the complex contrast operator in matrix form. $\Lambda$ converts the coefficients array $\bm{w}$ to a diagonal matrix. A typical choise of channel importance is $\alpha_L = 0.5, \alpha_{AB} = 1.5$. We will fix such setting for all the experiments.

\vspace{-6pt}
\paragraph{$\ell_2$-norm Solution:}

When $\ell = 2$, \eref{Eq:OBJ3} takes a quadratic form, which is minimized when:
\begin{equation}
	\begin{split}
		\frac{\partial E(\bm{g})}{ \partial \bm{g} } & = 0
		\\
		& = -\Lambda(\bm{w})(\bm{l} - \bm{g})
		\\
		& ~~~~ -\alpha_L C^\top (C \bm{l} - C\bm{g})
		-\alpha_{AB} C^\top (C \bm{a} - C\bm{g})
		-\alpha_{AB} C^\top (C \bm{b} - C\bm{g}),
	\end{split}
	\label{Eq:SOL1}
\end{equation}

This can be written as:
\vspace{-6pt}
\begin{equation}
	\vspace{-6pt}
	M \bm{g} = \bm{u},
	\label{Eq:SOL2}
\end{equation}
where:
\vspace{-6pt}
\begin{equation}
	\vspace{-6pt}
	M  = \Lambda(\bm{w}) + (\alpha_L + 2\alpha_{AB})C^\top C,
	\label{Eq:SOL3}
\end{equation}
and 
\vspace{-6pt}
\begin{equation}
	\vspace{-6pt}
	\bm{u}  = \Lambda(\bm{w}) \bm{l} + \alpha_L C^\top C \bm{l} + \alpha_{AB} C^\top C \bm{a} + \alpha_{AB} C^\top C \bm{b}.
	\label{Eq:SOL4}
\end{equation}

Notice that $M$ is invertable because:

1. $\Lambda(\bm{w})$ is a positive definite diagonal matrix.

2. $C^\top C$ is semi-positive definite.

3. $\alpha_L, \alpha_{AB}$ are positive.

1, 2, 3 together infer that $M$ is a positive-definite symmetric matrix and is thus invertable \cite{Lang02}. Therefore, the grayscale image is found by:
\vspace{-6pt}
\begin{equation}
	\vspace{-6pt}
	\bm{g} = M^{-1} \bm{u}.
	\label{Eq:SOL5}
\end{equation}

\paragraph{$\ell_1$-norm Solution:}
It is well known that $\ell_2$ norm based objective functions tend to over-smooth boundaries, this problem is significant for color-to-gray conversion.
This is because a grayscale image is intended to preserve opposing terms: brightness and multi-channel color contrast.
Unfortunately, it is unlikely to find a solution to suit the contrast for all three channels simultaneously. 
The $\ell_1$ solution allows us to neglect a relatively insignificant channel and strongly emphasize the contrast of a more significant channel.
In addition we take advantage of using $\ell_1$ where discontinuity at boundaries is implicitly handled in the graphical model.

To solve the $\ell_1$ objective function, we use the Iteratively Re-weighted Least Square method \cite{Candes08}.
For $i$-th iteration, we solve a reweighted least square problem of \eref{Eq:OBJ3}:
\vspace{-6pt}
\begin{equation}
	\vspace{-6pt}
	\begin{split}
		\min E(\bm{g}_i) &= \min \Lambda(\bm{q}_i)[E_B(\bm{d}) + E_C(\bm{g}_i)],
	\end{split}
	\label{Eq:SOL6}
\end{equation}
where the weight is defined as:
\vspace{-12pt}
\begin{equation}
	\vspace{-6pt}
	\bm{q}_{i+1} = \frac{1}{|\bm{r}_i| + 1},
	\label{Eq:SOL7}
\end{equation}
$w_0$ is set to 1 for all the pixels, and
$\bm{r}$ is the residual of each iteration:
\vspace{-6pt}
\begin{equation}
	\vspace{-6pt}
	\bm{r}_i = \Lambda(\bm{q}_i)[| \Lambda(\bm{w})(\bm{l} - \bm{g}_i) | + \alpha_L | C\bm{l} - C\bm{g}_i |
	+ \alpha_{AB} | C\bm{a} - C\bm{g}_i | + \alpha_{AB} | C\bm{b} - C\bm{g}_i |].
	\label{Eq:SOL8}
\end{equation}

\vspace{-6pt}
\paragraph{Speedup:}
Constructing the complex contrast matrix $C$ will yield a large dense matrix with $N^2$ elements where $N$ is the number of pixels, as a large scale DoG compares the difference over a large range. Fortunately, we find that the blurring matrix $C$ can be pre-computed using decomposition of the Gaussian kernel and the construction of an image pyramid. Details of the speedup can be found in the supplementary material.

\begin{figure}[tb]
	\includegraphics[width=\linewidth]{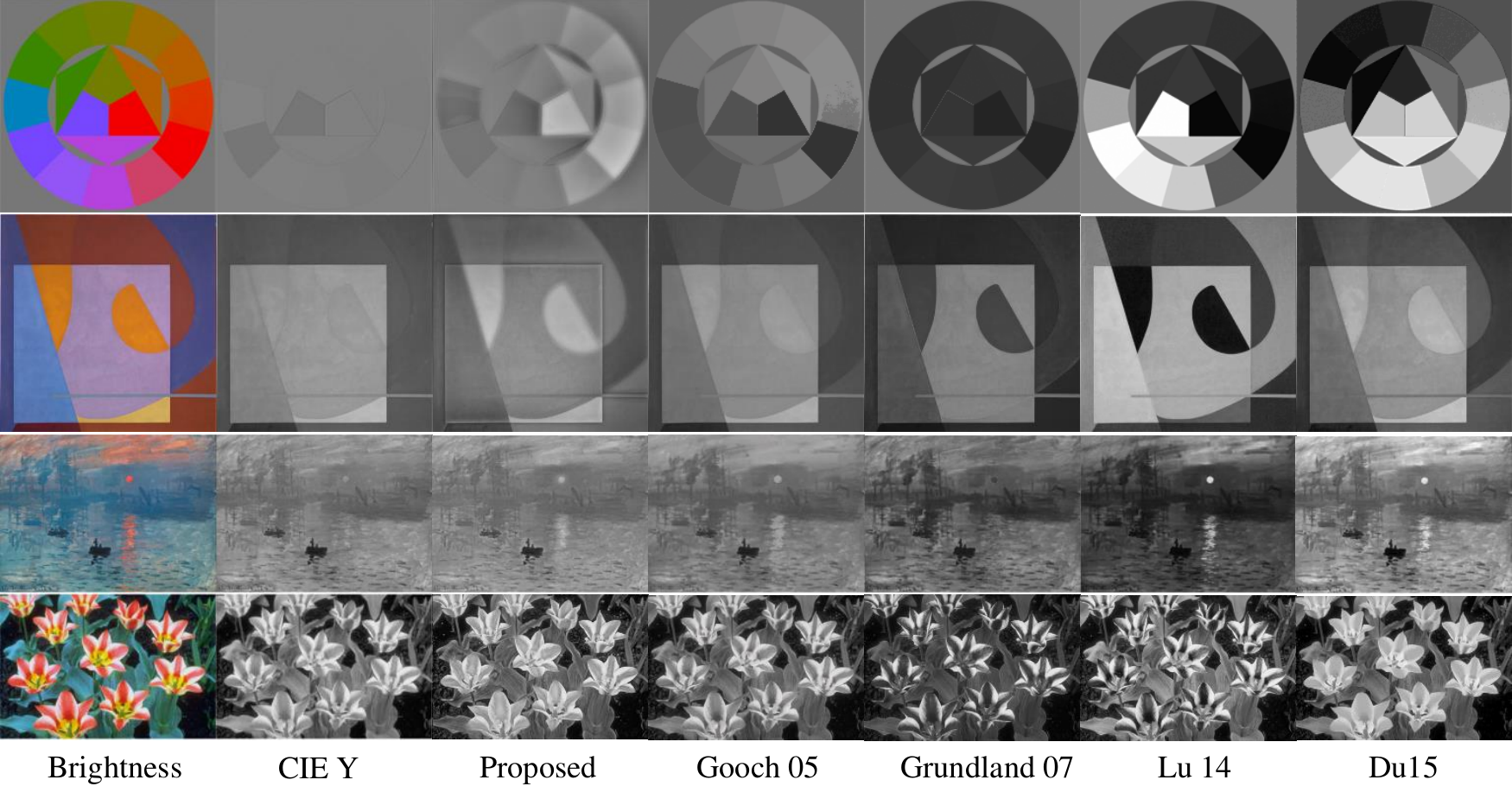}
	\vspace{-24pt}
	\caption{Visual results on \cadik dataset. (Optimized for A4 paper viewed from 40cm away).}
	\label{Fig:VisualCadik}
\end{figure}

\begin{figure}[tb]
	\vspace{-6pt}
	\includegraphics[width=\linewidth]{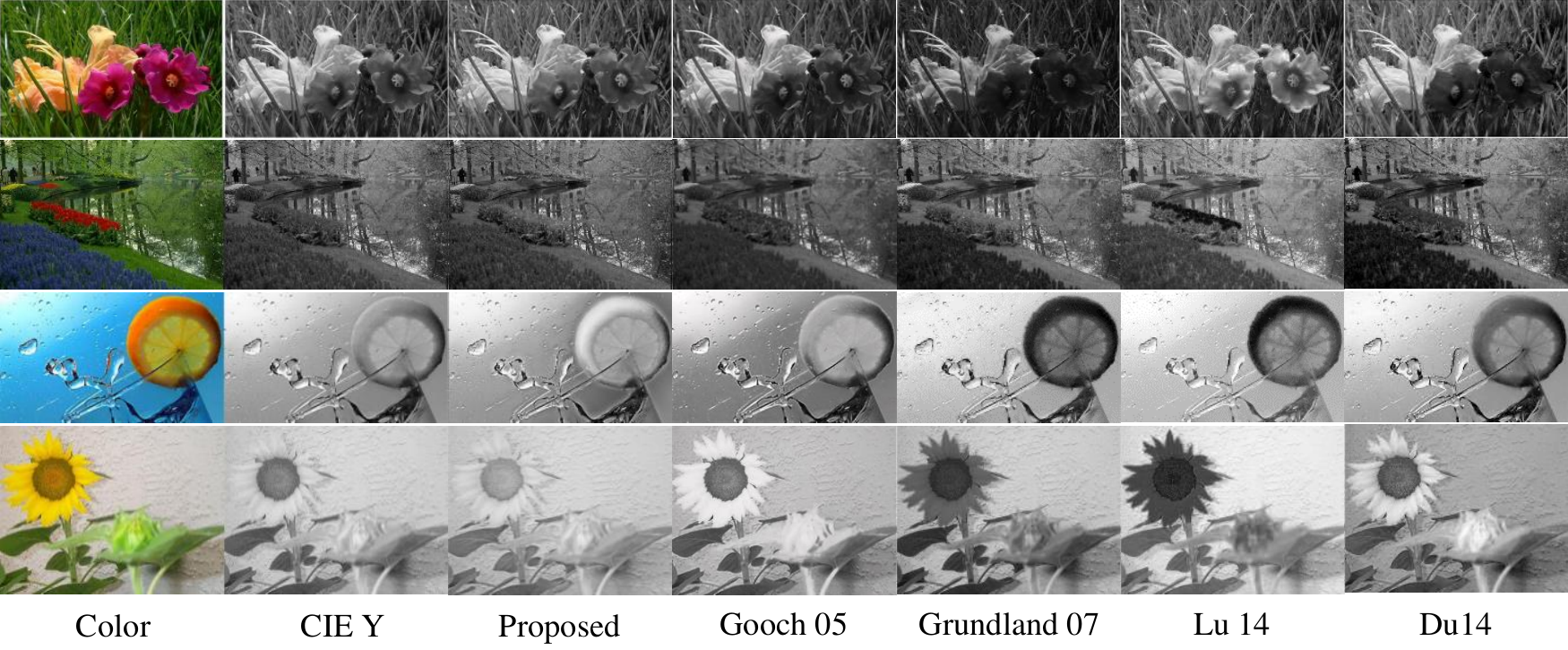}
	\vspace{-24pt}
	\caption{Visual results on Color250 dataset. (Optimized for A4 paper viewed from 40cm away).}
	\label{Fig:Visual250}
	\vspace{-12pt}
\end{figure}

\begin{figure}[htbp]
	\includegraphics[width=\linewidth]{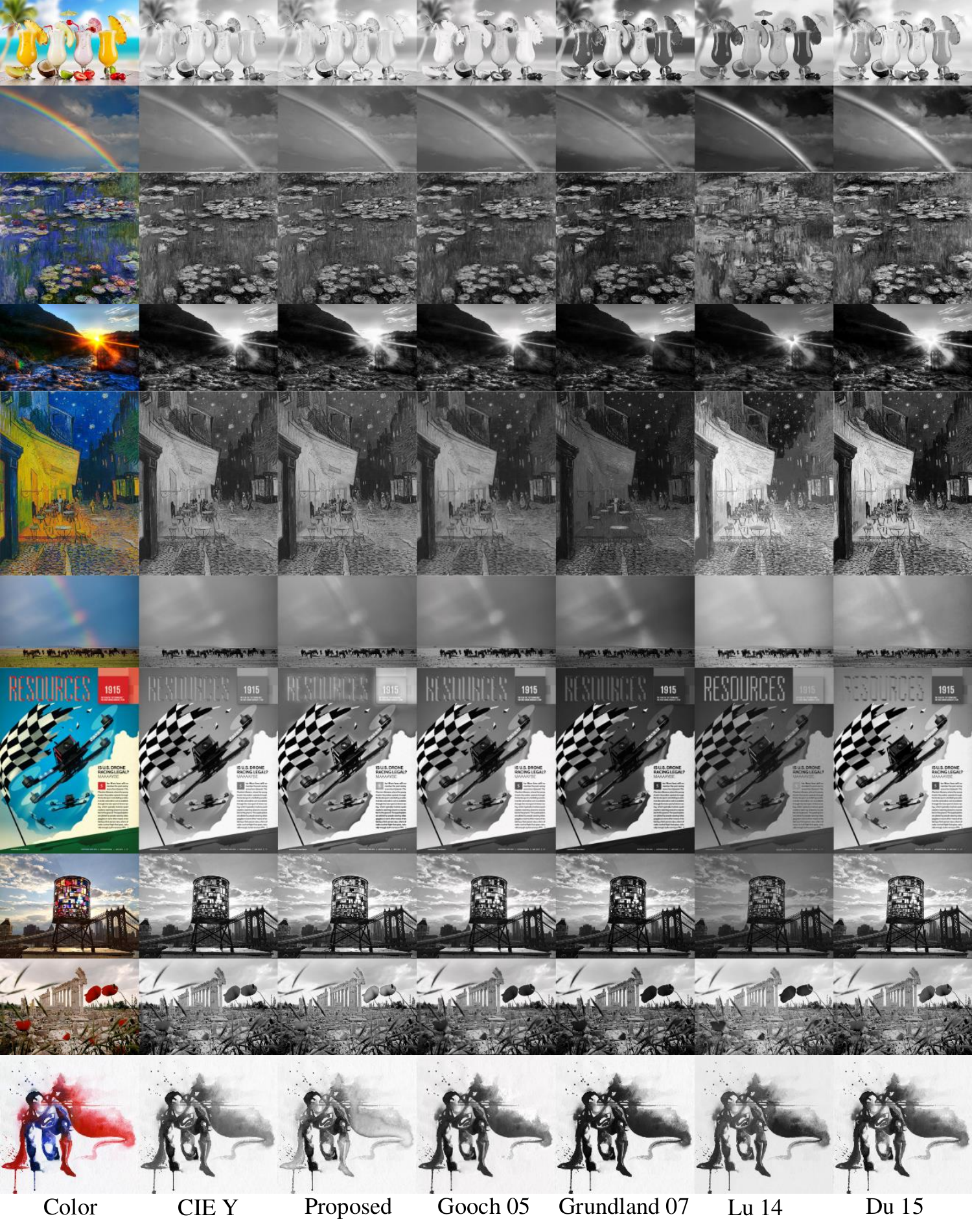}
	\caption{Visual results on NeoColor dataset. (Optimized for A4 paper viewed from 40cm away).}
	\label{Fig:VisualNeo}
\end{figure}

\section{Experiments}

We validate the proposed method systematically. 
We compare it with methods for which code are available, on existing datasets as well as the new proposed dataset.
We evaluate qualitative performance, as well as performance with respect to metrics.
Further, we include a user study to validate the method and the proposed metric.

\subsection{Datasets}
\vspace{-6pt}
{\em \cadik}
\cite{Cadik08} is the earliest dataset for C2G evaluation, which contains 25 highly saturated images. The contents varies from geometric pattern to real scenes.
Because the \cadik dataset has only 25 images, as is common practice, we evaluate the performance on the {\em Color250} dataset\cite{Lu14}.
This data set is a subset of the 2001 Berkeley Segmentation Dataset \cite{Martin01}.
{\em CSDD}
\cite{Du15} contains 22 highly saturated and fully chromatic geometric patterns. Because the number of images is limited and no natural image are included, we do not use it in our evaluation. However, we include the results on the CSDD dataset in the supplementary material.

\vspace{-6pt}
\paragraph{NeoColor}
The Berkeley Segmentation Dataset was designed for evaluating segmentation performance rather than evaluating C2G algorithms.
Images in the dataset tend to have strong contrast on object boundaries, only contain a limited number of segments with different colors, and the contents are relatively simple.
On the other hand, C2G algorithms must be able to deal with more complex images that frequently arise.

To fully evaluate the performance of C2G algorithms, we have collected high quality digital images of natural scenes, advertisements, computer graphic designs as well as fine art. 
We consider that this new dataset, NeoColor, extends upon the existing datasets by including images of greater color complexity.
A full collection of the dataset appears in the supplementary material.
Some examples are shown in \fref{Fig:VisualNeo}.

\vspace{-6pt}
\subsection{Evaluation metric}

We evaluate our methods using {\em EScore} proposed by Lu \etal
\cite{Lu14} which has a joint evaluation between color contrast preservation and color fidelity. 
However, the metric emphasizes preservation of high contrast, and measurement of spatial contrast preservation is based on random sampling which does not fully reflect the concept on visual perception preservation.
The {\em QScore} Ma \etal
\cite{Ma15}, has better quantization for color contrast measurement, however, scale dependent contrast is not emphasized.

\vspace{-9pt}
\paragraph{Visual Perception Metric (PScore)}
Because both metrics focus more on contrast preservation rather than the overall perception of color, we provide further evaluation using the proposed perceptually consistent color energy function \eref{Eq:OBJ1}. For a fair comparison, we weight all scales of contrast preservation separately, we set $\beta_i = 1$ (\eref{Eq:CCM1}) for the evaluated scale and rest as 0.

\vspace{-6pt}
\subsection{Results}
\vspace{-6pt}
We compare with classic methods of Gooch \etal \cite{Gooch05} and Grundland \etal \cite{Grundland07} as well as the state-of-the-art methods of Lu \etal \cite{Lu14} and Du \etal \cite{Du15}. We use the default setting for all the datasets. For our method, scale is set to optimize contrast for A4 paper viewed from 40cm away, and all parameters are fixed for all the experiments. 
Code is not available\footnote{Not available on webpage as well as through email contact} for the methods of Liu \etal \cite{Liu15} and Ji \etal \cite{Ji15} and thus they are not evaluated.
We report both visual and quantitative results. 

\vspace{-6pt}
\paragraph{Visual results}
Visual results on the \cadik dataset, the Color250 dataset and NeoColor dataset are shown in \fref{Fig:VisualCadik}, \fref{Fig:Visual250} and \fref{Fig:VisualNeo} respectively.
A full collection of the results can be found in the supplementary material.

As can be seen, methods that do not consider perceptual consistency are likely to produce strong contrast but unnatural images.
For example in \fref{Fig:VisualNeo}, black orange juice, a rainbow with black and white stripes, and a black sun in a sunset image, all lose their natural impression.
Further, the impressionist painting "Lotus" which now has more emphasis on weeds has lost the original feeling of the image, and the intention of the artist.
On the contrary, the proposed methods allows a reasonable sense of color contrast as well as the original feeling of the images.

\begin{figure}[tb]
	\includegraphics[width=\linewidth]{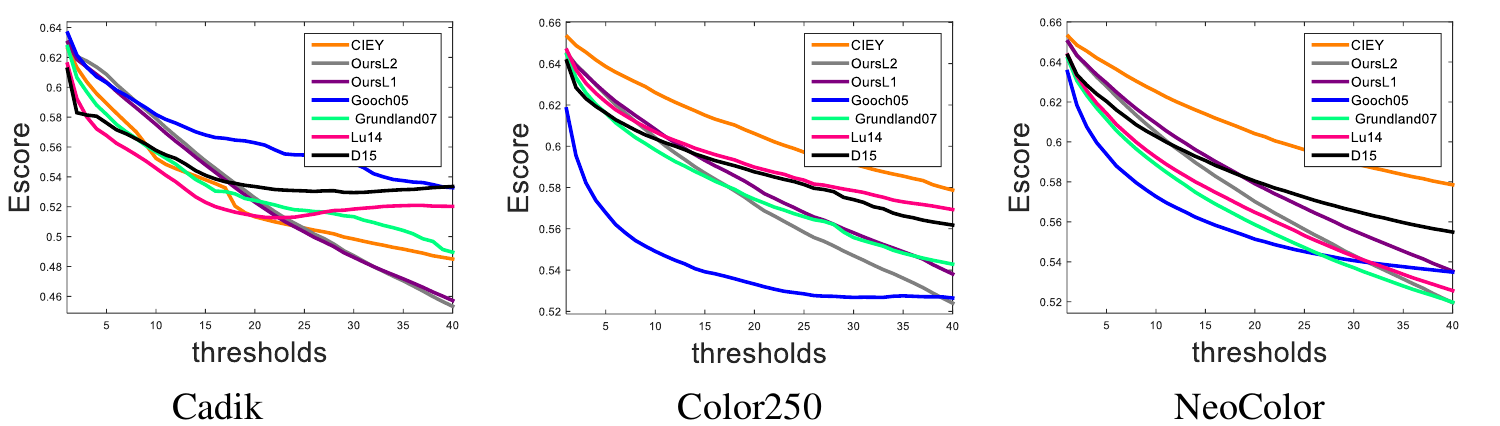}
	\vspace{-24pt}
	\caption{Quantitative evaluation using the EScore \cite{Lu14}.} The y-axis the Escore, where as the x-axis the contrast threshold level.
	\label{Fig:EScore}
	\vspace{-12pt}
\end{figure}
\begin{figure}[tb]
	\includegraphics[width=\linewidth]{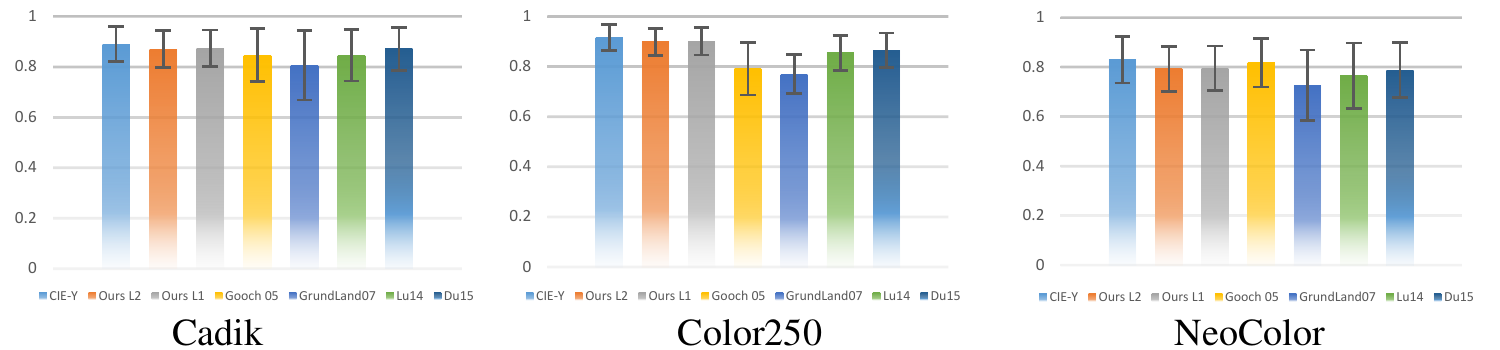}
	\vspace{-24pt}
	\caption{Quantitative evaluation using the QScore \cite{Ma15}. Higher score is better. Average score and standard variance is shown.}
	\vspace{-0pt}
	\label{Fig:QScore}
\end{figure}
\begin{figure}[tb]
	\vspace{-6pt}
	\includegraphics[width=\linewidth]{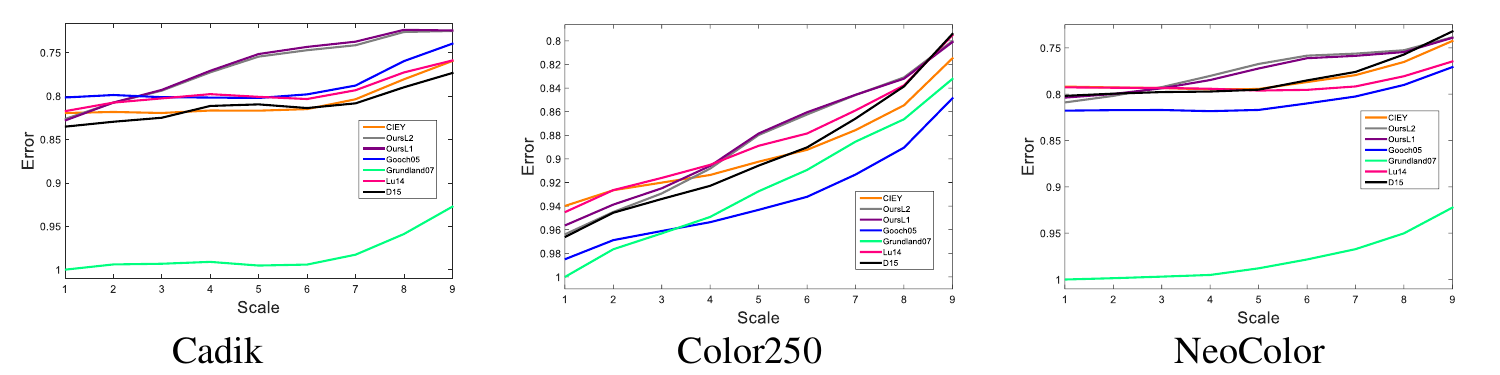}
	\vspace{-24pt}
	\caption{Quantitative evaluation using proposed PScore. The y-axis is the PScore. The x-axis represents the natural contrast preservation at scale $2 ^ {\frac{i + 2}{2}}$ pixels.}
	\label{Fig:PScore}
	\vspace{-12pt}
\end{figure}

\vspace{-6pt}
\paragraph{Quantitative results:}
Quantitative evaluation using the E-score is 
shown in \fref{Fig:EScore}, a higher curve indicates better performance. As can be seen, the \cadik dataset is not large enough to produce a smooth curve. On the Color250 dataset, our method performs well on contrast in the range 0-20 (where total black to white was a scale of 100).

The Q-score proposed by Ma \etal \cite{Ma15} is shown in \fref{Fig:QScore}.
Whereas the evaluation for most C2G algorithms was reasonable, CIE Y which does not consider color contrast is evaluated as the best on all the datasets.

Results for consistency (P-Score) are shown in \fref{Fig:PScore}.
Again, a higher curve shows better performance. 
The x-axis represents contrast preservation at scale $2 ^ {\frac{i + 2}{2}}$ pixels.
We plot the contrast preservation for different scales. As can be seen, the proposed method best preserves the color contrast at the scale which is most sensitive for perception. Whereas for the scales which are less contributive to overall contrast, the proposed method has lower contrast preservation in comparing with other methods.
The ordering is accordance with the feeling on visual results.

\vspace{-6pt}
\subsection{User Study}
Ten volunteers (aged 22-41) with normal or corrected-to-normal vision took part, with data collection in March 2016. Written consent was obtained from participants before they began the experiment. The research was approved by the {\em  institution removed} Ethics Committee, and adhered to the tenets of the Declaration of Helsinki.

Considering the complexity of the task, four methods were compared: the proposed, CIE Y (as a widely used method), Lu \etal (as the state-of-the-art), Grundland \etal (as reported relatively worse from the quantitative metric).
Participants were masked to the algorithm.
The color image was presented in the middle of the screen, with a white background, and the four grayscale images presented at identical distances from the original image. Please refer to supplementary material for screen layout and experimental details.
We selected 50 hard images from Color250, and 200 images from NeoColor, excluding more difficult images. Then, 150 images were randomly selected from these subsets with a 30\% selection probability from Color250 and a 70\% probability from NeoColor.
Each user was shown these images in random order, and placement of the grayscale images was randomly assigned for each trial.

Participants were required to rank the grayscale images from best (1) to worst (4) according to the instruction: ``Which black-and-white image best represents the original color image". When answering, the participants were asked to consider: which grayscale image best retains the contrast between colors of the original color image; and, which grayscale image best retains the overall sense or feeling of the original color image.
The chance rate of ranking a image as "best" (1 and 2) representation of the color image was 50\%, and 75\% was the criterion set as a benchmark as a reliable grayscale representation of the color image (\ie, participants ranking a Color2Gray vision processing method ``1" or ``2" for 75\% of the trials).
Descriptive analyses were used to characterize the counts and percentages of responses. Comparisons between participants and color-to-gray methods for the average rankings of the preferred images were calculated using non-parametric statistics (e.g., Kruskal-Wallis H test). Windows SPSS v23 (IBM Corporation, Somers, NY) was used for all statistical analyses.

\begin{figure}[tb]
	\includegraphics[width=\linewidth]{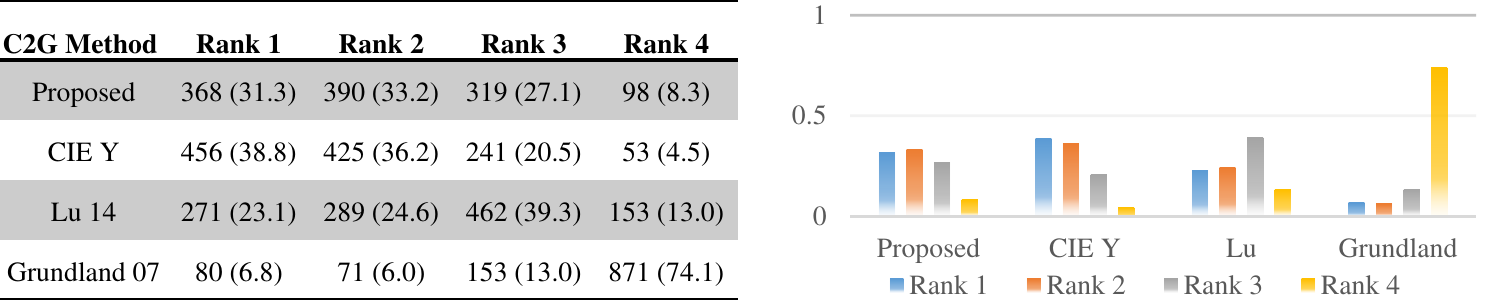}
	\vspace{-18pt}
	\caption{User preference on C2G algorithms.} Left, statics of users' ranking by number of choices. Parenthesis list the percentage. Right, ranking frequency distribution of all the methods.
	\vspace{-12pt}
	\label{Fig:UserStudy}
\end{figure}

\vspace{-6pt}
\paragraph{Results:}
The data for ten participants was pooled for further analyses as there was no significant difference in overall preferences between participants ($p < 1.00$).
Overall preference is shown in \fref{Fig:UserStudy}.
Our proposed C2G method was ranked as a significantly ($p<0.0001$) better representation of the original color images compared to the Grundland and Lu  methods. However, the CIE-Y approach was selected significantly ($p<0.0001$) more often compared to the proposed method. 
Our proposed method was ranked as being a better (ranked 1 or 2) representation of the original color image for 64.5\% of responses, and CIE-Y achieved 75.0\% of responses ranked as 1 or 2.
The CIE-Y approach is similar to approaches that are ubiquitous in print media (\eg, newspapers, kindle).
Among the specialized C2G algorithms, our proposed method was highest ranked.
The Grundland and Lu methods were consistently ranked as the least preferred methods with only 12.8\% and 47.7\% of responses ranked 1 or 2.

\section{Conclusion}
\vspace{-12pt}
In this paper, we proposed a method for converting color images to gray scale while preserving brightness consistency as much as possible. 
The key ideas of brightness consistency is that the feeling and naturalness of the original color is preserved, and that color contrast is preserved.
These key ideas were used to derive quantitative metrics based on recent studies in vision science.
An $\ell_1$ optimization framework was proposed to find the grayscale image which optimised the proposed brightness consistency metric. 
To evaluate the proposed methods, we used both existing datasets and a proposed new dataset. 
We also validated both the algorithm and the metric with a user study.
For future work, we will exploit brightness consistency with more emphasis on visual augmentation.
\clearpage

\bibliographystyle{splncs}
\bibliography{YSDbib}
\end{document}